\documentclass{article}

\usepackage[preprint]{corl_2026} 
\usepackage{graphicx}
\usepackage{caption}
\usepackage{algorithm}
\usepackage{algorithmic}
\usepackage{amsmath,amssymb}
\usepackage{booktabs} 
\usepackage{multirow}
\usepackage{wrapfig}
\usepackage{float}

\usepackage{xcolor} 
\title{HaWMPO: Hallucination-Aware World Model-based Policy Optimization for Generalist Robot Policy}

\author{
    Zengjue Chen$^{1,2}$ \quad
    Peidong Liu$^{1,\dagger}$ \quad
    Jiawei Li$^{1}$ \quad
    Qi Wang$^{2,\dagger}$ \\[6pt]
    $^{1}$Joy Future Academy, JD \\
    $^{2}$School of Artificial Intelligence, Jilin University \\[4pt]
    \texttt{zengjue24@mails.jlu.edu.cn,
    qiwang@jlu.edu.cn,
    lpd19@tsinghua.org.cn} \\[4pt]
    {\normalfont\footnotesize
    $^{\dagger}$Corresponding authors.}
}
\begin{document}
\maketitle


\begin{abstract}
 Generalist robot policies have demonstrated strong generalization across robotic manipulation tasks, yet their success rates remain limited in complex long-horizon scenarios. Recent methods improve Visual-Language-Action (VLA) policies through online reinforcement learning on real robots, but such training relies on costly physical interactions, suffers from low sample efficiency, and may introduce hardware and safety risks. World models offer a promising alternative by enabling policy optimization with imagined rollouts. However, long-horizon rollouts generated by world models often suffer from prediction hallucinations, producing biased state transitions that can mislead policy learning. To address this issue, we propose Hallucination-aware World Model-based Policy Optimization (HaWMPO), a closed-loop reinforcement learning pipeline for VLA policy post-training with world models. Specifically, HaWMPO introduces an action-conditioned hallucination-aware model to estimate the reliability of generated image sequences, and incorporates hallucination scores into group relative policy optimization through a Reward-Soft mechanism, suppressing unreliable action chunks during training. On the LIBERO benchmark, HaWMPO achieves the best average success rate, with gains of 15.0\% over the base model and 2.8\% over the strongest baseline; real-world experiments on a G1 robot further validate its effectiveness, raising the average success rate on two manipulation tasks from 67.5\% to 80.0\%.

\end{abstract}

\keywords{Reinforcement Training, World Model, Generalist Robot Policy} 


\section{Introduction}
\label{sec:intro}
In recent years, generalist robot policies\cite{kim2024openvla,black2024pi_0,brohan2023rt,kim2025fine,wu2026pragmatic,shi2025memoryvla,rt1,shukor2025smolvla,Nvidia2025GR00TNA,Luo2026BeingH05SH,xu2026reaction,Academy2026JoyAIRA0A,Team2026JoyAIRA0S,Liu2026JoyAISimAS} have demonstrated strong multi-task generalization capabilities in robotic manipulation tasks. However, their further improvement is still constrained by the high cost of interaction in real-world environments. This limitation is particularly evident in embodied manipulation tasks, where directly performing online reinforcement learning on physical robots often requires extensive physical trial and error. Such a process is not only sample-inefficient, but also introduces potential risks of hardware damage and safety issues. Therefore, a key challenge for improving general-purpose robotic policies is how to provide closed-loop feedback for reinforcement learning while reducing the reliance on real-world robot interactions.
\begin{figure}[htbp]
  \centering
  \includegraphics[width=1\textwidth]{ 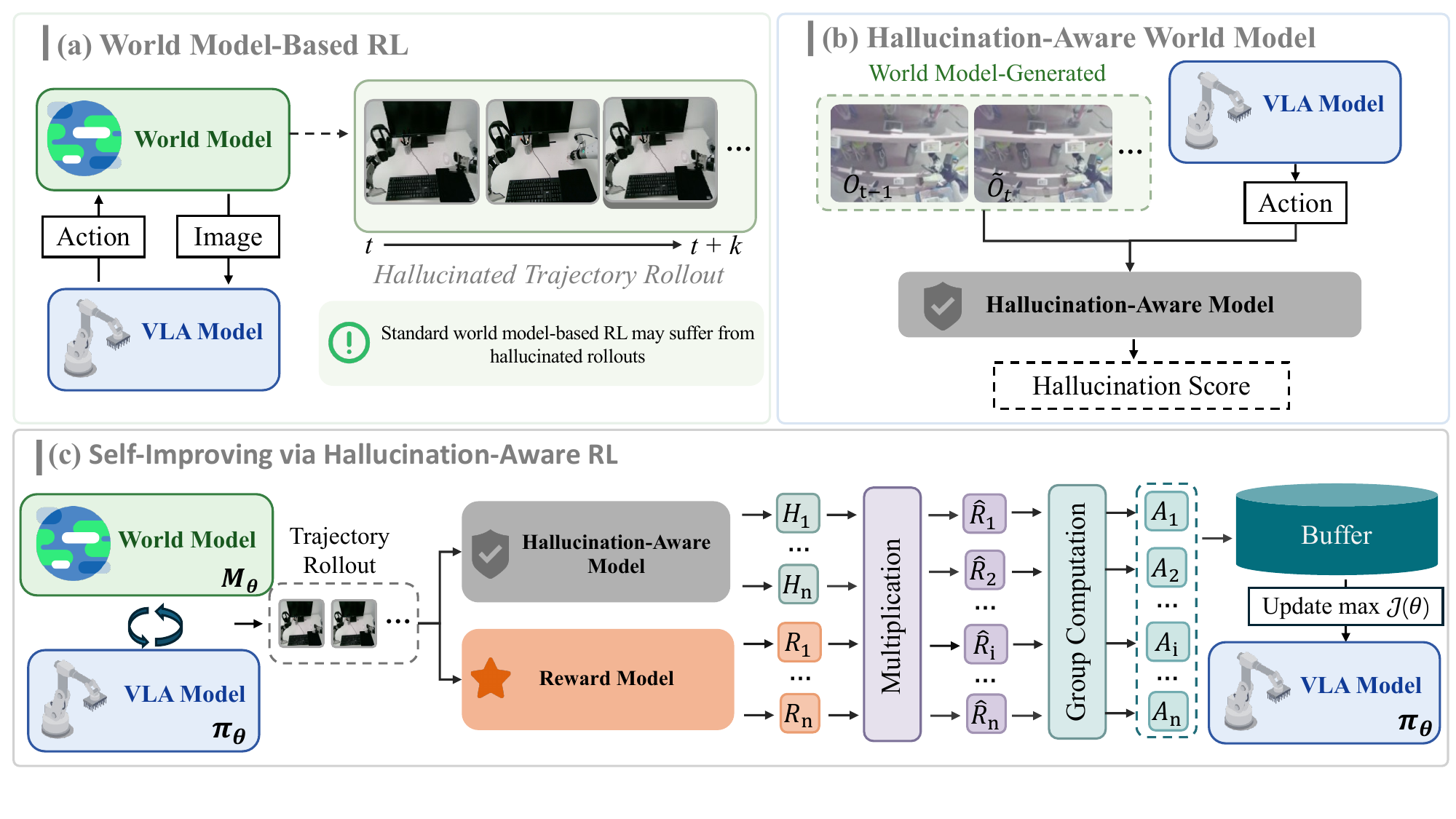}
  \caption{Illustration of the proposed hallucination‑aware VLA reinforcement learning pipeline. \textbf{(a)} In standard world model-based RL, the VLA policy interacts with a learned world model to generate imagined trajectory rollouts, while hallucinated visual predictions may introduce biased reward signals and lead to unstable policy updates. \textbf{(b)} Conditioned on VLA actions, the hallucination-aware model assesses the reliability of imagined trajectory rollouts from the world model and assigns a hallucination score to each rollout. \textbf{(c)} Our closed‑loop pipeline incorporates hallucination scores into GRPO with the Reward‑Soft mechanism to stabilize policy optimization.}
  \label{fig:overall}
  \vspace{-10pt}
\end{figure}
With the rapid development of world models\cite{wan2025wan,ali2025world,zhu2024sora,Kim2026CosmosPF}, transferring reinforcement learning from the real environment to an imagination space constructed by world models has gradually emerged as a promising paradigm for policy optimization. In this work, a world model can serve as an interactive virtual simulator, providing low-cost and efficient synthetic experience for policy learning. However, generative world models are not strict physical simulators. Their predictions are inherently based on generative modeling, which makes them prone to accumulated prediction errors and hallucinations during long-horizon rollouts. Directly using hallucinated virtual trajectories for reinforcement learning may mislead policy updates, causing the VLA policy to exploit world-model biases rather than learn actions effective in real environments.
For example, WoVR\cite{jiang2026wovr} shows that hallucinated imagined rollouts can mislead policy optimization by turning world-model errors into false learning signals. Such hallucinations introduce biased state transitions, causing the policy optimization process to deviate from real-world dynamics and ultimately weakening the stability of reinforcement learning and the performance of the learned policy.

We propose a hallucination-aware world model-based reinforcement learning pipeline for VLA policy optimization. Instead of treating all world-model-generated rollouts equally, our method explicitly estimates their reliability and weights their contributions according to the predicted hallucination level. The pipeline consists of three core components: an action-conditioned world model that generates future image sequences from the current visual context and VLA-predicted action chunks, a hallucination-aware model that evaluates rollout reliability under the given action condition, and a reward model that predicts task completion from generated images. Building on this world model-based RL pipeline, we introduce a Reward-Soft mechanism to reduce the negative influence of highly hallucinated action chunks during reward shaping and policy updates, thereby improving training stability and policy performance. We validate our method in the LIBERO\cite{liu2023libero} simulation benchmark and real-world robotic settings, focusing on task success rates, training stability, and deployment performance. Overall, the main contributions of this work are summarized as follows:
\begingroup
\setlength{\leftmargini}{1em}  
\begin{itemize}
\item We establish a closed-loop VLA reinforcement learning pipeline integrating a world model, a hallucination-aware model, and a reward model. This architecture supports online-style policy optimization without frequent real-world robot interactions.

\item We design an action-conditioned hallucination-aware model to quantify the reliability of image sequences predicted by the world model under specified action inputs. This model produces a continuous hallucination score that can be directly leveraged for reinforcement learning.

\item We integrate the derived hallucination score into the GRPO optimization pipeline, and propose a Reward-Soft mechanism to penalize severely hallucinated action segments during rollouts. This mitigates the adverse bias of the world model on policy updates in reinforcement learning.

\end{itemize}
\endgroup

\section{Related Work}
\label{sec:Related Work}
\subsection{Reinforcement Learning for VLA Models}
As VLA models advance toward universal robotic policies, research increasingly adopts reinforcement learning\cite{chen2026last,chen2025conrft,guo2025improving,zhang2024grape,Yu2025RLinfFA,Yuan2025SceneR1VL} to break limits of supervised imitation learning. Traditional imitation learning depends heavily on offline demonstrations, with poor performance on out-of-distribution cases, long-horizon tasks and sparse rewards. VLA-RL\cite{lu2025vla} adopts a robotic process reward model to mitigate sparse reward issues and optimizes policies via online PPO\cite{schulman2017proximal} reinforcement learning. Targeting flow-matching VLA models, the $\pi$RL\cite{chen2025pirl} framework devises Flow-Noise and Flow-SDE to resolve intractable action log-likelihood calculation, enabling direct RL optimization and boosting flow-based policy capacity. LWD\cite{wang2026learning} constructs an offline-online RL framework for robot clusters. With distributional implicit value learning and Q-function adjoint matching, it addresses several limitations of pre-trained VLA models, including over-dependence on imitation learning, real-world distribution drift, sparse rewards in long-horizon tasks, underutilized failure data, and limited cross-task generalization across diverse robotic manipulation scenarios.

\subsection{World Models for VLA Reinforcement Training} 

With the development of VLA models, world models have become an important tool for VLA training due to their ability to simulate environment distributions and capture physical dynamics. Recent works have incorporated world models into VLA pre-training, post-training, and reinforcement learning to reduce real-world interaction costs and provide virtual experience. For example, World-Env\cite{ali2025world} enables RL post-training without real-world interaction by combining a geometry-aware world simulator with a VLM-guided instant reflector. WMPO\cite{zhu2025wmpo} integrates a pixel-level video world model with online GRPO\cite{shao2024deepseekmath}, improving VLA policies through policy-behavior alignment and frame-level action control. However, long-horizon autoregressive generation in world models can accumulate errors and lead to hallucinations. To mitigate this issue, WoVR\cite{jiang2026wovr} introduces Keyframe Initialization and Refinement (KIR) and PACE-based policy-world model co-evolution to reduce hallucinations, error accumulation, and distorted RL signals caused by policy drift.

\section{Methodology}
\label{sec:method}

\subsection{Hallucination-Aware Model for World Model}

World-model prediction errors can accumulate during imagined rollouts
and introduce misleading signals into VLA policy optimization.
We introduce an action-conditioned Hallucination-Aware Model (HAM)
to estimate the reliability of each generated video chunk.
HAM takes an eight-frame predicted chunk $\hat{\mathbf{V}}_c$,
its conditioning actions $\mathbf{a}_c$, the current observation $I_c$,
and an anchor image $I_0$ as inputs. Here, $I_0$ is the initial
observation of the rollout, providing a fixed visual reference.

To construct supervision, we generate video chunks from recorded
actions in successful and failed trajectories and compare them with
the corresponding observed sequences. We use four quality-oriented
metrics: DINOv3 similarity, depth consistency, optical-flow-based
trajectory consistency, and MUSIQ image quality. Each metric
$s_c^{(m)}$ is min--max normalized, and the hallucination target is
defined as
\begin{equation}
    q_c^{(m)} =
    \operatorname{clip}_{[0,1]}
    \left(
        \frac{s_c^{(m)}-\ell_m}
        {\max(u_m-\ell_m,\varepsilon)}
    \right),
    \qquad
    h_c^{\star} =
    1-\frac{1}{4}\sum_{m=1}^{4}q_c^{(m)},
    \label{eq:ham_target}
\end{equation}
where $\ell_m$ and $u_m$ are the normalization bounds and
$\varepsilon>0$ ensures numerical stability. All metrics are oriented
so that higher values indicate better quality; thus, larger
$h_c^{\star}$ indicates more severe hallucination.
Depth and trajectory consistency capture geometric and motion
discrepancies beyond visual appearance, although this composite
target remains a proxy rather than a guarantee of physical validity.
Task failure alone does not imply hallucination.

\begin{figure}[!t]
    \centering
    \includegraphics[width=0.95\textwidth]{ 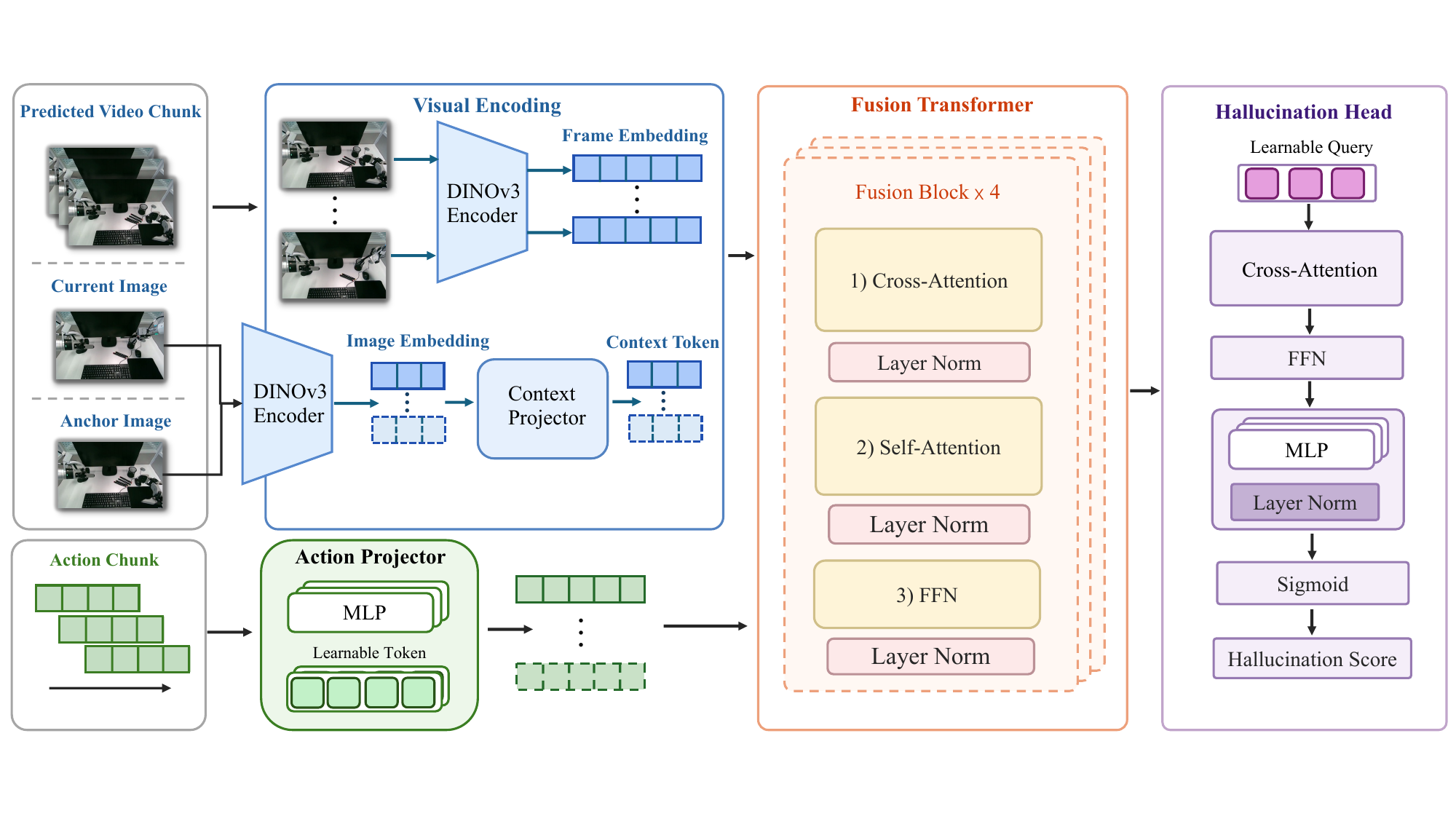}
    \caption{Architecture of HAM. Predicted video frames, the current
    observation, the initial anchor image, and conditioning actions
    are encoded and fused to predict a chunk-level hallucination
    score. Higher scores indicate less reliable predictions.}
    \label{fig:overall}
    \vspace{-10pt}
\end{figure}

As illustrated in Fig.~\ref{fig:overall}, a frozen DINOv3 encoder
extracts visual features, while the Context and Action Projectors
map contextual images and actions into a shared feature space.
A four-layer Fusion Transformer combines these inputs through
action-to-visual cross-attention, self-attention, and feed-forward
layers. A learnable query aggregates the fused representations,
followed by an MLP and sigmoid activation to produce a continuous
chunk-level score. We train HAM using mean squared error:
\begin{equation}
    \hat{h}_c =
    f_{\psi}(\hat{\mathbf{V}}_c,\mathbf{a}_c,I_c,I_0),
    \qquad
    \mathcal{L}_{\mathrm{HAM}} =
    \frac{1}{|\mathcal{D}_{\mathrm{HAM}}|}
    \sum_{c\in\mathcal{D}_{\mathrm{HAM}}}
    (\hat{h}_c-h_c^{\star})^2,
    \label{eq:ham_loss}
\end{equation}
where $\mathcal{D}_{\mathrm{HAM}}$ is the labeled training set.
Observed future sequences are required only for target construction.
During imagined rollouts, HAM predicts $\hat{h}_c$ from the generated
chunk and its conditioning inputs for subsequent reward modulation.

\subsection{Reward‑Soft Mechanism for Hallucination‑Aware RL }
After training the aforementioned hallucination‑aware model, we introduce the quantitative hallucination metric into the reinforcement learning pipeline and further propose the Reward‑soft mechanism accordingly. Specifically, this mechanism imposes hallucination‑driven penalties on seemingly successful RL steps. We first average the hallucination scores within each action chunk sampled in the current rollout, then transform the raw reward via Equation (1) to achieve dynamic reward modulation conditioned on hallucination severity.
\begin{equation}
\hat{R} = (1 - \alpha \cdot H) \cdot R
\end{equation}

where $\hat{R}$ denotes the reward modulated by the Reward‑soft mechanism, $H$ represents the average hallucination score of the current action chunk predicted by the hallucination‑aware model, $R$ is the average raw reward provided by the reward model for the corresponding chunk, and $\alpha$ is a tunable hyperparameter that governs the impact of overall hallucination level on the reward. We verify that the Reward‑soft mechanism significantly mitigates the negative effects of high‑hallucination RL steps on the optimization of the Vision‑Language‑Action (VLA) model. In the subsequent experimental section, we further analyze the influence of hyperparameter $\alpha$ on the performance of the Reward‑soft mechanism.
\begin{algorithm}[t]
\caption{Hallucination‑Aware World Model-Based Policy Optimization (HaWMPO)}

\textbf{Require:} Pretrained VLA policy $\pi_\theta$; action‑conditioned world model $M_\phi$;hallucination‑aware model $H$; reward model $R$; synthetic rollout budget $N$;parallel env number $G=8$; iterations $K_\text{iter}$; hallucination penalty hyperparameter $\alpha$

\textbf{Output:} Post‑trained policy $\pi_\theta$
\begin{algorithmic}[1]
\STATE Initialize synthetic imaginative environments from $M_\phi$
\FOR{$i=1$ to $K_\text{iter}$}
    \STATE \textbf{(1) Parallel Synthetic Rollouts in World Model}
    \STATE \quad Roll out $\pi_\theta$ in $G$ parallel imaginative environments of $M_\phi$
    \STATE \quad Collect action‑chunk‑level trajectories $\tau_1,\dots,\tau_N$

    \STATE \textbf{(2) Hallucination Score Estimation}
    \STATE \quad Compute average hallucination score $H_c$ for each action chunk via $H$

    \STATE \textbf{(3) Reward‑Soft Modulated Reward Calculation}
    \STATE \quad Compute raw reward $R_c$ for each chunk via reward model $R$
    \STATE \quad Modulate reward by $\hat{R}_c = (1-\alpha\cdot H_c)\cdot R_c$ (Eq. 1)

    \STATE \textbf{(4) GRPO‑based VLA Policy Post‑training}
    \STATE \quad Calculate group‑wise normalized advantages from modulated rewards $\hat{R}_c$
    \STATE \quad Update $\pi_\theta$ by minimizing GRPO loss with hallucination‑aware rewards
\ENDFOR
\STATE \textbf{return} $\pi_\theta$
\end{algorithmic}
\end{algorithm}
\subsection{Hallucination-Aware World Model-Based Policy Optimization}
Subsequently, we formally propose the Hallucination‑Aware World Model Policy Optimization method. By integrating the Vision‑Language‑Action (VLA) model, action‑conditioned world model, hallucination‑aware model and reward model, we adopt the GRPO algorithm improved by the Reward‑Soft mechanism to conduct online reinforcement learning training for VLA agents. 
\begin{gather}
    {A}_{i,t} = \frac{{\hat{R}_{i,t}}-\frac{1}{N}\sum_{i=1}^{N} \hat{R}_{i,t}}{\sqrt{\frac{1}{N-1} \sum_{i=1}^{N} (\hat{R}_{i,t} - \frac{1}{N}\sum_{i=1}^{N} \hat{R}_{i,t})^2})}.
\end{gather}
As illustrated in Figure 1, the world model serves as a simulator, enabling the VLA agent to perform online RL within the imaginative space constructed by the world model. Specifically, the action‑conditioned world model takes the current frame, initial reference frame, and action chunk output by the VLA model as inputs to generate the subsequent 8 future frames. 
\begin{gather}
\rho_{i,t} = \frac{\pi_{\theta}(a_{i,t} \mid s_{i,t})}{\pi_{\theta_{\text{old}}}(a_{i,t} \mid s_{i,t})}.
\end{gather}
The VLA model then conducts rollout for the next action chunk based on these generated frames. Meanwhile, the reward model produces raw rewards for each action chunk from the synthesized visual observations, while the hallucination‑aware model outputs the average hallucination score of the corresponding chunk. The Reward‑Soft mechanism is further applied to derive hallucination‑modulated optimized rewards. Built upon the GRPO framework, our overall training pipeline leverages 8 parallel imaginative environments constructed by the world model, where the VLA agent executes rollouts concurrently across all environments. 

\begin{equation}
\begin{aligned}
\mathcal{J}_{\text{HaWMPO}}(\theta) 
&= \mathbb{E}\!\left[q \sim P(Q), \{o_i\}_{i=1}^N \sim \pi_{\theta_{\mathrm{old}}}(O \mid q)\right] \cdot \frac{1}{N} \sum_{i=1}^{N} \frac{1}{|o_i|} \sum_{t=1}^{|o_i|} \\
&\quad \cdot \Biggl\{ \min \Biggl[ \rho_{i,t} \text{Adv}_{i,t},\operatorname{clip}\!\left(\rho_{i,t},1-\epsilon,1+\epsilon \right) \text{Adv}_{i,t} \Biggr] - \beta D_{\mathrm{KL}}\!\bigl[\pi_\theta \parallel \pi_{\text{ref}}\bigr] \Biggr\}.
\end{aligned}
\end{equation}

Using the optimized rewards obtained via Reward‑Soft adjustment, we compute relative advantages through group‑wise reward normalization, and finally formulate the GRPO loss to update the VLA model end‑to‑end.
\begin{equation}
D_{\mathrm{KL}}\!\left[\pi_\theta \parallel \pi_{\text{ref}}\right] = \frac{\pi_{\text{ref}}(a_{i,t} \mid s_{i,t})}{\pi_{\theta}(a_{i,t} \mid s_{i,t})} - \log \frac{\pi_{\text{ref}}(a_{i,t} \mid s_{i,t})}{\pi_{\theta}(a_{i,t} \mid s_{i,t})} - 1.
\end{equation}

\section{Experiments}
\label{sec:experiments}
\subsection{Experimental Settings}
In this experiment, we follow the experimental setup of prior work and adopt OpenVLA-OFT as the base model. Before reinforcement learning, the base model is first supervised fine-tuned using a one-shot SFT setting, where only one expert trajectory is used for each task to obtain an initial policy. To ensure a fair comparison with existing methods, we continue to use the LIBERO simulation benchmark as the primary evaluation environment. In our pipeline, the world model serves as a simulator for the LIBERO environment, while the VLA policy is optimized through reinforcement learning within the virtual interaction environment constructed by the world model. Training and evaluation are mainly conducted on three LIBERO task suites: Object, Spatial, and Goal. We train the model using the Adam optimizer. The learning rate of the backbone network is set to $2.0 \times 10^{-5}$, while the learning rate of the value network is set to $3.0 \times 10^{-3}$. The weight decay coefficient is set to 0.01. Gradient clipping is applied during training, with the maximum gradient norm constrained to 1.0. All experiments are conducted on 8 NVIDIA H200 GPUs.

\subsection{Simulation Experiments}

We first conduct simulation experiments on the LIBERO benchmark to evaluate the effectiveness of the proposed hallucination-aware reinforcement learning pipeline. We compare HaWMPO with three representative baselines: the vanilla OpenVLA-OFT policy, WMPO, and WoVR*. All methods are evaluated on three LIBERO task suites (Spatial, Goal, and Object), and we report the success rate on each suite together with the average.

It is worth noting that our reproduced WoVR* baseline does not adopt the original PACE strategy used in the WoVR paper. Instead, for a fair comparison under the same experimental protocol, all world-model-based methods in our experiments train the world model only once and then use it as a fixed simulator for VLA policy optimization. Moreover, all VLA policies are trained with reinforcement learning for 200 steps before evaluation. Therefore, the results of WoVR* reported in this paper may differ from those in the original WoVR paper, as they correspond to our reproduced setting rather than the full original training pipeline.

\begin{table}[H]
  \centering
  \small
  \setlength{\tabcolsep}{14pt}
  \renewcommand{\arraystretch}{0.95}
  \caption{Comparison of success rates across Spatial, Goal, and Object task categories. Our method achieves the highest average performance against baseline approaches.}
  \label{tab:main_results}
  \vspace{2pt}
  \begin{tabular}{lcccc}
    \toprule
    Method & Spatial & Goal & Object & Avg \\
    \midrule
    OpenVLA-OFT-base & 61.5 & 48.2 & 36.3 & 48.7 \\
    WMPO             & 67.8 & 54.6 & 48.0 & 56.8 \\
    WoVR*            & 69.2 & \textbf{64.0} & 49.6 & 60.9 \\
    \textbf{HaWMPO (Ours)}    & \textbf{77.2} & 61.6 & \textbf{52.2} & \textbf{63.7} \\
    \bottomrule
  \end{tabular}
  \vspace{-6pt}
\end{table}

As shown in Table~\ref{tab:main_results}, HaWMPO achieves the best average success rate of 63.7\%, improving over the vanilla OpenVLA-OFT policy by 15.0 points, over WMPO by 6.9 points, and over the strongest baseline WoVR* by 2.8 points. On individual suites, HaWMPO achieves the best results on Spatial (77.2\%) and Object (52.2\%), surpassing WoVR* by 8.0 and 2.6 points respectively, and remains comparable to WoVR* on Goal. These results demonstrate that reinforcement learning in the world-model-based imagination space can substantially improve the task execution capability of the VLA policy, and that hallucinated world-model rollouts can bias VLA policy learning. By explicitly modeling rollout hallucination and integrating it via the Reward-Soft mechanism, our method reduces the impact of unreliable synthetic trajectories, enabling more robust policy optimization and stronger LIBERO performance.

\subsection{Real-World Experiments}

\noindent
\begin{minipage}[t]{0.49\linewidth}
    \vspace{0pt}

    We evaluate HaWMPO on the G1 robot using two tasks
    (Fig.~\ref{fig:real_world_setup}):
    \textit{Tissue-to-Box}, placing a tissue inside a white storage
    box, and \textit{Headphone-on-Stand}, placing headphones onto
    a stand. Success requires the tissue to remain inside the box
    or the headphones to remain supported by the stand after release.

    \par\medskip

    We compare against the SFT-trained OpenVLA-OFT base policy and
    WoVR, with 20 trials per method per task under the same evaluation
    protocol. Table~\ref{tab:real_world} shows that HaWMPO achieves
    the highest observed success rates on both tasks: 85\% and 75\%.
    Its average success rate of 80.0\% exceeds the base policy and
    WoVR by 12.5 and 7.5 percentage points, respectively.
    These results provide preliminary evidence of benefits in
    physical manipulation; broader evaluation remains necessary.

\end{minipage}
\hfill
\begin{minipage}[t]{0.48\linewidth}
    \vspace{0pt}
    \centering

    \begin{minipage}[t]{0.48\linewidth}
        \vspace{0pt}
        \centering
        \parbox[c][2.1cm][c]{\linewidth}{%
            \centering
            \includegraphics[
                width=\linewidth,
                height=2.1cm,
                keepaspectratio
            ]{ 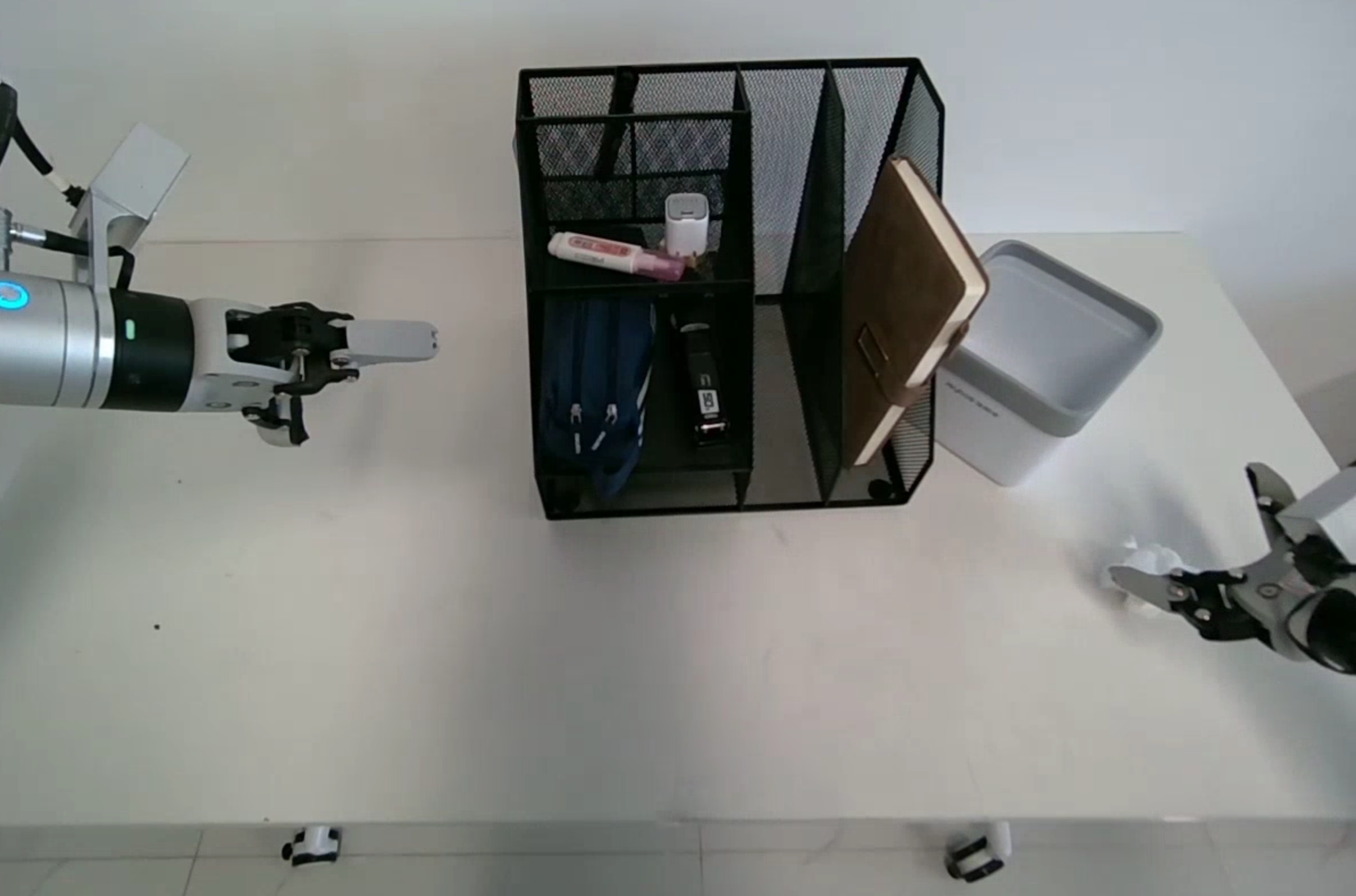}%
        }
        \par\vspace{2pt}
        {\footnotesize (a) Tissue-to-Box\par}
    \end{minipage}
    \hfill
    \begin{minipage}[t]{0.48\linewidth}
        \vspace{0pt}
        \centering
        \parbox[c][2.1cm][c]{\linewidth}{%
            \centering
            \includegraphics[
                width=\linewidth,
                height=2.1cm,
                keepaspectratio
            ]{ 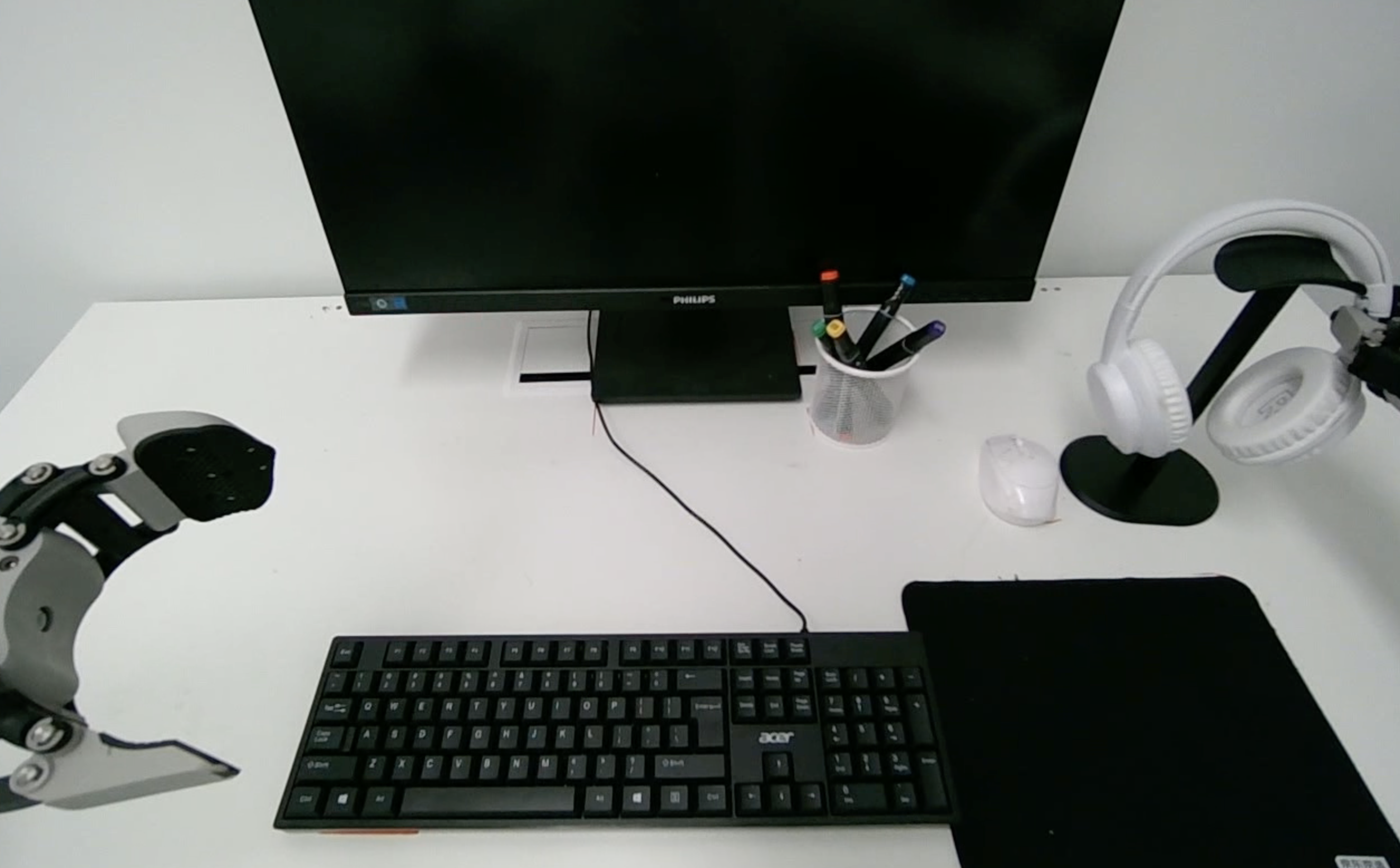}%
        }
        \par\vspace{2pt}
        {\footnotesize (b) Headphone\par}
    \end{minipage}
    \par

    \begingroup
        \captionsetup{skip=4pt}
        \captionof{figure}{Real-world task setups on the G1 robot.}
        \label{fig:real_world_setup}
    \endgroup

    \vspace{8pt}

    \begingroup
        \captionsetup{position=top,skip=4pt}
        \captionof{table}{Success rates (\%); successes out of
        20 trials in parentheses.}
        \label{tab:real_world}

        \footnotesize
        \setlength{\tabcolsep}{2pt}
        \renewcommand{\arraystretch}{1.10}
        \begin{tabular*}{\linewidth}
            {@{\extracolsep{\fill}}lcc@{}}
            \toprule
            Method & Tissue-to-Box & Headphone \\
            \midrule
            Base VLA & 75\% (15/20) & 60\% (12/20) \\
            WoVR     & 80\% (16/20) & 65\% (13/20) \\
            \textbf{HaWMPO}
                & \textbf{85\% (17/20)}
                & \textbf{75\% (15/20)} \\
            \bottomrule
        \end{tabular*}
        \par
    \endgroup

\end{minipage}
\par\medskip

\subsection{Analysis of the Reward-Soft Mechanism}

We further investigate the effect of the Reward-Soft mechanism, focusing on the hyperparameter $\alpha$ in Eq.~(1), which controls how strongly the hallucination score modulates the final reward.

\begin{figure}[t]
  \centering
  \includegraphics[width=\textwidth]{ 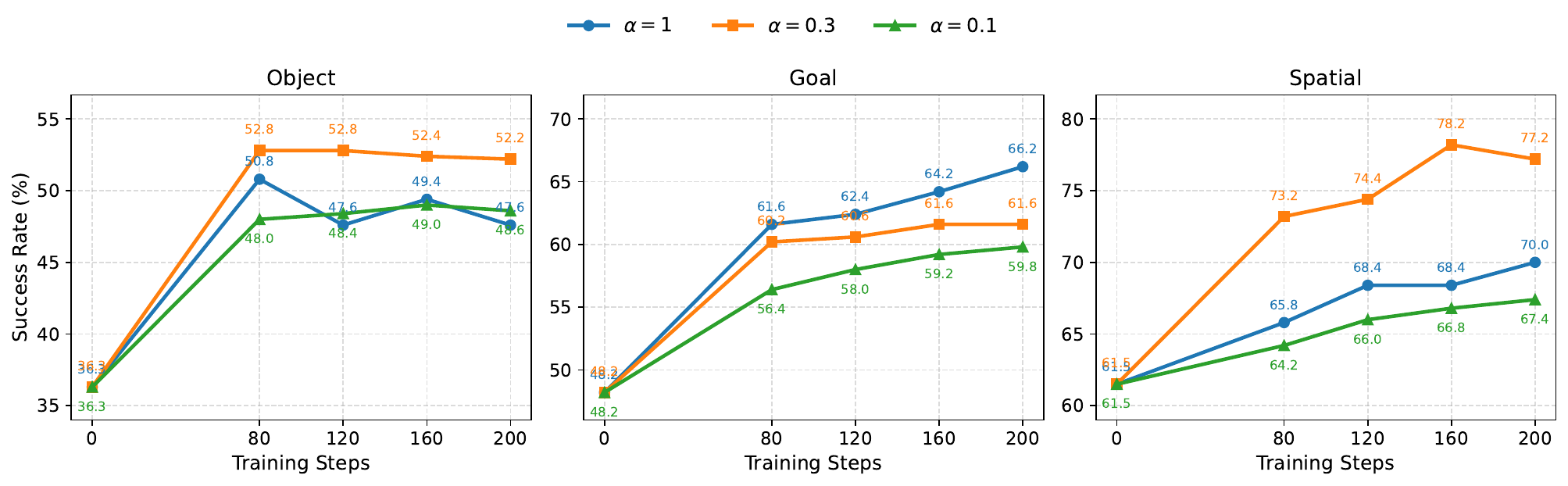}
  \caption{Ablation study on $\alpha$ across three LIBERO task suites. A moderate penalty ($\alpha=0.3$) is preferred on Object and Spatial, while Goal benefits from a stronger penalty ($\alpha=1$).}
  \label{fig:alpha}
\end{figure}

Intuitively, an overly large $\alpha$ suppresses the reward of high-hallucination rollouts too aggressively, discarding virtual experience that remains useful despite visual imperfections; an overly small $\alpha$ instead lets unreliable trajectories bias policy updates. As shown in Fig.~\ref{fig:alpha}, the optimal penalty strength is task-dependent: $\alpha=0.3$ achieves the best and most stable results on Object and Spatial (52.0\% and 77.2\% at 200 steps), whereas Goal, whose rollouts suffer more severe hallucination accumulation, benefits from a stronger penalty ($\alpha=1$, 66.2\%). In all cases, $\alpha=0.1$ improves the slowest, confirming that hallucination-aware reward modulation is essential. We therefore select $\alpha$ per task suite and report the corresponding results in Table~1.

\subsection{Analysis of the Hallucination-Aware Model}

We evaluate HAM independently of downstream policy performance
to assess its reliability signal for Reward-Soft. The evaluation
uses 50 video chunks from 50 trajectories in LIBERO
, covering 10 tasks with five chunks per task.
Human annotations identify 18 hallucinated and
32 non-hallucinated chunks.

We compare HAM with four single-metric baselines: MUSIQ image
quality, DINO feature similarity, optical-flow trajectory
consistency, and depth discrepancy. All scores are oriented
so that higher values indicate greater predicted hallucination.
Each method is evaluated on the same annotated chunks using
AUROC and average precision (AP), with 95\% confidence intervals
estimated from 2,000 trajectory-level bootstrap resamples.
DINO, trajectory, and depth baselines require corresponding
observed future frames as references; HAM and MUSIQ do not.

\begin{table}[!htbp]
    \centering
    \small
    \setlength{\tabcolsep}{14pt}
    \renewcommand{\arraystretch}{1.12}

    \caption{Hallucination detection against human annotations
    on 50 LIBERO chunks. Brackets show trajectory-bootstrap
    95\% confidence intervals.
    $^\dagger$Requires observed future frames.}
    \label{tab:ham_single_metric}

    \begin{tabular}{lcc}
        \toprule
        Method & AUROC $\uparrow$ & AP $\uparrow$ \\
        \midrule
        MUSIQ-only
            & 0.3056 [0.1410, 0.4798]
            & 0.3750 [0.2079, 0.5561] \\
        DINO-only$^\dagger$
            & 0.9062 [0.8182, 0.9733]
            & 0.8299 [0.6638, 0.9563] \\
        Trajectory-only$^\dagger$
            & 0.8872 [0.7883, 0.9610]
            & 0.8069 [0.6276, 0.9388] \\
        Depth-only$^\dagger$
            & 0.8941 [0.7912, 0.9745]
            & 0.8644 [0.7258, 0.9649] \\
        \textbf{HAM}
            & \textbf{0.9375} [0.8648, 0.9911]
            & \textbf{0.8952} [0.7660, 0.9885] \\
        \bottomrule
    \end{tabular}
\end{table}
As shown in Table~\ref{tab:ham_single_metric}, HAM achieves
the highest AUROC (0.9375) and AP (0.8952).
The reference-based metrics also provide informative signals,
whereas MUSIQ-only performs substantially worse, suggesting
that image quality alone is insufficient for detecting
hallucinations in this evaluation set. HAM provides this
reliability signal without requiring observed future frames, making it applicable during imagined rollouts.
These comparisons are descriptive; statistical significance of the differences has not been established.
Separately, HAM achieves an MAE of 0.0174, an RMSE of 0.0250, and a Spearman correlation of 0.9310 against the composite proxy targets. These regression metrics measure agreement with the constructed supervision, whereas the human-label evaluation independently assesses hallucination detection. Fig.~\ref{fig:hmcase} provides complementary qualitative examples: higher HAM scores correspond to more evident discrepancies from the observed frames. Together, these findings support using HAM as a reliability signal for downweighting unreliable imagined chunks through Reward-Soft, complementing the downstream policy ablation.

\begin{figure}[!htbp]
    \centering
    \includegraphics[width=0.95\linewidth]{ 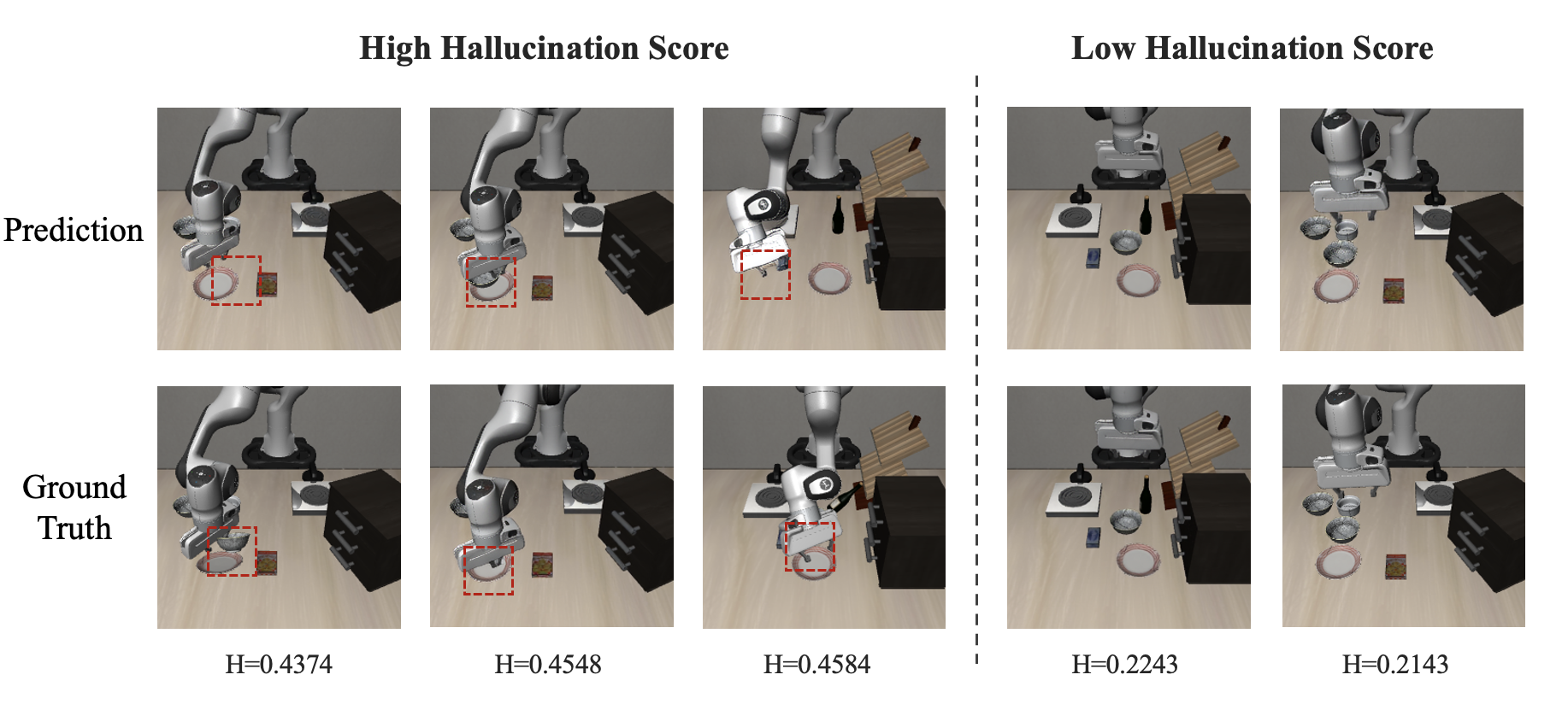}
    \caption{World-model predictions with higher and lower
    HAM scores. Higher-scored examples exhibit more evident
    discrepancies from the corresponding observed frames,
    whereas lower-scored examples more closely match
    the observations.}
    \label{fig:hmcase}
\end{figure}

\subsection{Ablation Study}

We further conduct an ablation study on the hallucination-aware model (HAM). Note that removing HAM not only eliminates the explicit estimation of hallucination levels in world-model-generated rollouts, but also disables the hallucination-score-based Reward-Soft mechanism during training. Therefore, HAM and Reward-Soft are treated as an integrated module and ablated together.

\begin{table}[H]
  \centering
  \small
  \setlength{\tabcolsep}{10pt}
  \renewcommand{\arraystretch}{0.95}
  \caption{Ablation study of the hallucination-aware model on Spatial, Object, and Goal task suites. $\Delta$ denotes the improvement of our method over the variant without the hallucination-aware model.}
  \label{tab:hallucination_ablation}
  \vspace{2pt}
  \begin{tabular}{lcccc}
    \toprule
    Task Suite & Steps & w/o HAM & Ours & $\Delta$ \\
    \midrule
    \multirow{3}{*}{Spatial}
    & 80  & 61.6 & \textbf{73.2} & +11.6 \\
    & 120 & 64.6 & \textbf{74.4} & +9.8 \\
    & 200 & 69.2 & \textbf{77.2} & +8.0 \\
    \midrule
    \multirow{3}{*}{Object}
    & 80  & 50.4 & \textbf{52.8} & +2.4 \\
    & 120 & \textbf{54.2} & 52.8 & -1.4 \\
    & 200 & 49.6 & \textbf{52.2} & +2.6 \\
    \midrule
    \multirow{3}{*}{Goal}
    & 80  & \textbf{60.4} & 60.2 & -0.2 \\
    & 120 & \textbf{62.4} & 60.6 & -1.8 \\
    & 200 & \textbf{64.0} & 61.6 & -2.4 \\
    \bottomrule
  \end{tabular}
  \vspace{-6pt}
\end{table}

As shown in Table~\ref{tab:hallucination_ablation}, removing HAM leads to a clear performance drop on Spatial and Object. The degradation is most pronounced on Spatial, where our full model improves the success rate by 8.0--11.6 points at all evaluated steps, and the gain remains as large as 8.0 points at 200 training steps, indicating that the benefit of hallucination-aware training persists rather than vanishing as the policy converges. On Object, our method also achieves higher success rates at most training steps.

On Goal, the full model under the unified $\alpha=0.3$ setting falls slightly behind the ablated variant, though the gap stays within 2.4 points at all evaluated steps. We emphasize that this stems from the choice of penalty strength rather than the ineffectiveness of HAM: as shown in Fig.~\ref{fig:alpha}, Goal benefits from a stronger hallucination penalty, and with $\alpha=1$ our method reaches 66.2\% on Goal, surpassing the ablated variant (64.0\%). For fairness, we fix $\alpha=0.3$ across all task suites instead of tuning it per task. Overall, these results indicate that explicitly modeling world-model hallucinations and incorporating them into policy optimization via the Reward-Soft mechanism improves the reliability of virtual rollouts and reduces the negative impact of low-quality generated trajectories on policy updates.
\section{Conclusion}
\label{sec:conclusion}
In this work, we propose a hallucination-aware world model-based reinforcement learning pipeline for Vision-Language-Action (VLA) policy optimization. Specifically, we design an action-conditioned hallucination-aware model to explicitly estimate the reliability of world-model-generated rollouts, and further incorporate the predicted hallucination scores into GRPO through the proposed Reward-Soft mechanism. By reducing the influence of unreliable synthetic trajectories during policy optimization, our method improves the stability and effectiveness of world-model-based reinforcement learning. Experimental results on LIBERO benchmarks demonstrate that the proposed framework consistently outperforms existing baselines across multiple task suites, verifying the importance of explicitly modeling rollout hallucinations for VLA policy learning.
\section{Limitations}
\label{sec:Limitations}
Despite the promising results, this work still has several limitations. First, the hallucination-aware model relies on manually designed supervision signals, including feature discrepancy, depth inconsistency, and trajectory deviation, which may not fully capture all types of world-model hallucinations. Second, the current framework is mainly evaluated on relatively short-horizon manipulation tasks in the LIBERO benchmark, and its scalability to more complex long-horizon real-world scenarios remains underexplored. In addition, the hallucination-aware model and world model are trained separately, which may limit the overall adaptability of the system during online reinforcement learning. Future work will investigate end-to-end joint optimization and more generalizable hallucination estimation strategies for large-scale embodied learning.



\bibliography{main}  

\clearpage

\appendix
\section*{Appendix}
\section{Implementation Details}
\subsection{Data Collection}
When training the hallucination-aware model, we first adopt the world model to perform chunk-wise inference and generation on the collected real dataset, so as to construct the predicted dataset. Next, we revise several metrics in WorldArena. Leveraging models including DINOv3 and Depth-Anything-v3 together with trajectory optical flow methods, we extract features from both real and predicted datasets. Four hallucination metrics corresponding to each chunk are calculated, namely depth information, DINO feature similarity, image quality and trajectory optical flow similarity. The resulting data is finally used as the training set for our hallucination-aware model.

\subsection{Training Data Analysis}
In robotic settings, world model hallucinations consist of intertwined failures in geometry, visual appearance, motion coherence and semantic alignment. We adopt four complementary metrics: Depth Accuracy for geometric consistency against ground truth, Image Quality scored by MUSIQ, Trajectory Accuracy for optical flow motion consistency via NDTW, and DINO Similarity for frame-level semantic and appearance matching using DINOv3.
\begin{figure}[!htbp]
  \centering
  \includegraphics[width=0.95\textwidth]{ 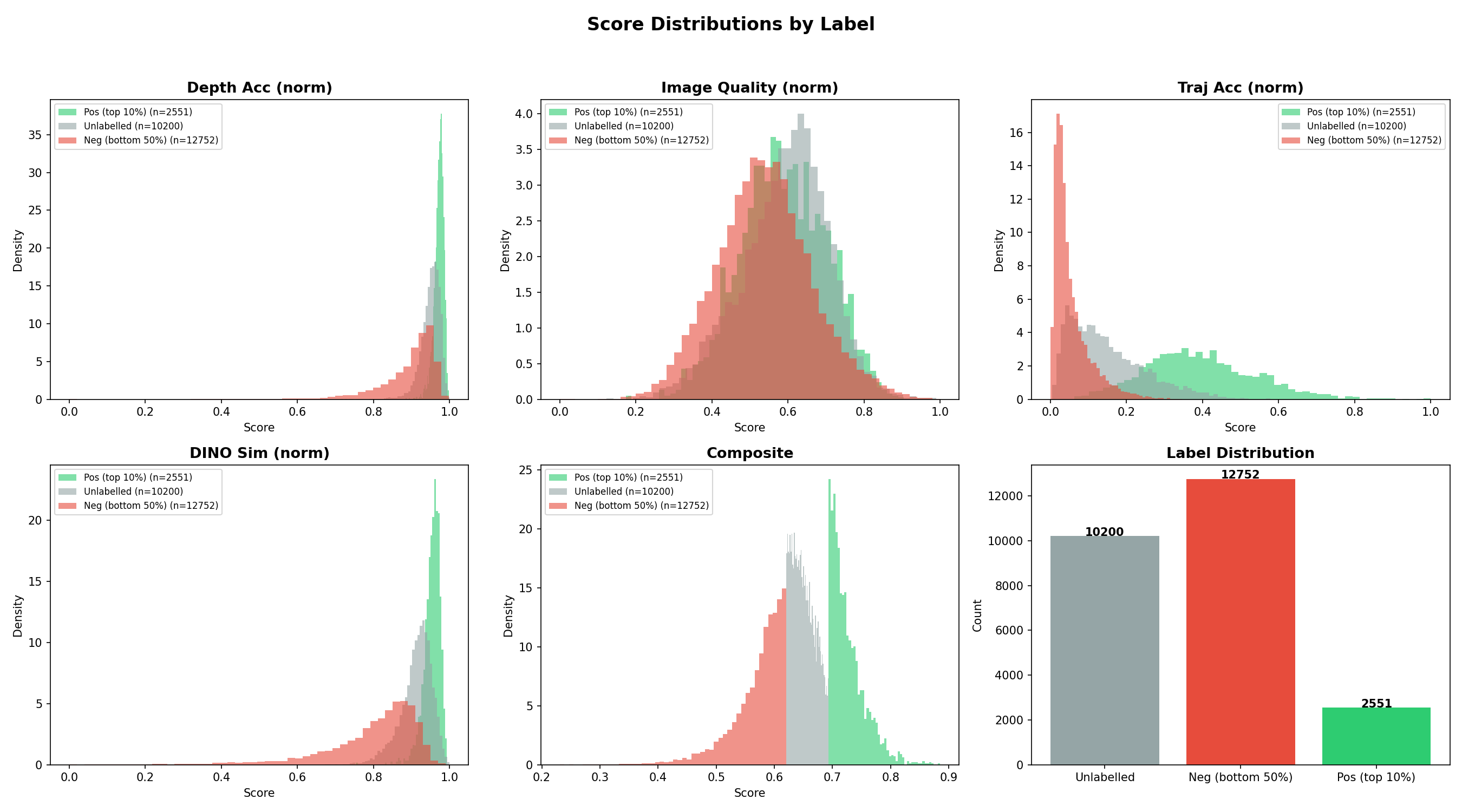}
  \caption{Score distributions of world-model prediction chunks across four normalized metrics and the composite score, grouped by Pos (top 10\%), Unlabelled (middle 40\%), and Neg (bottom 50\%). Trajectory accuracy shows the clearest separation between positive and negative samples; depth accuracy and DINO similarity are generally high but Neg exhibits a longer left tail; image quality distributions overlap heavily across groups; composite scores form three nearly non-overlapping bands, consistent with percentile-based labeling.
}
  \label{fig:hmcase}
\end{figure}
All metrics are min-max normalized and equally weighted to compute the Composite score. We split samples by scores: the top 10\% as high-quality anchors, the bottom 50\% as typical hallucination cases, and the remaining 40\% as unlabeled transitions. Statistics show Composite is strongly correlated with Trajectory Accuracy (0.72) and DINO Similarity (0.80), while Depth Accuracy provides auxiliary geometric constraints. Image Quality has nearly no correlation and even weak negative correlation with other metrics, proving good visual quality does not indicate the absence of hallucinations. PCA results further verify that motion and semantics dominate the data distribution with clear separability between positive and negative samples.

Instead of binary labels, we use continuous Composite scores as supervision. Taking 8-frame predictions, action sequences and contexts as input, the model regresses the deviation from real scenarios. It effectively detects critical failures in RL rollout such as abnormal motion, object drift and semantic inconsistency, without being affected by superficial visual features like sharpness.
\begin{figure}[!htbp]
  \centering
  \includegraphics[width=0.95\textwidth]{ 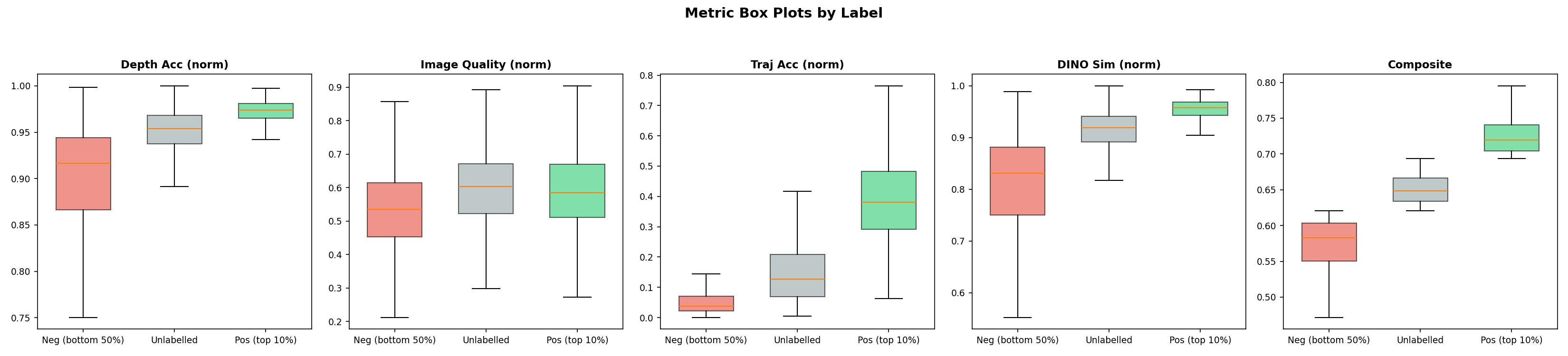}
  \caption{Median and interquartile ranges of each metric across the three label groups. Trajectory accuracy, DINO similarity, depth accuracy, and composite score increase monotonically from Neg to Pos, with composite showing near-complete separation of IQRs; image quality differs little across groups and is a poor standalone discriminator between faithful predictions and hallucinations.}
  \label{fig:hmcase}
\end{figure}
Given the heavily left-skewed Trajectory distribution, limited positive samples and score saturation of Depth and DINO metrics, we use Composite as a multi-metric proxy label for training to align hallucination detection with robot decision-making. Motion-related hallucinations should be prioritized, and image quality should not be used to judge hallucinations.

\begin{figure}[!htbp]
  \centering
  \includegraphics[width=0.95\textwidth]{ 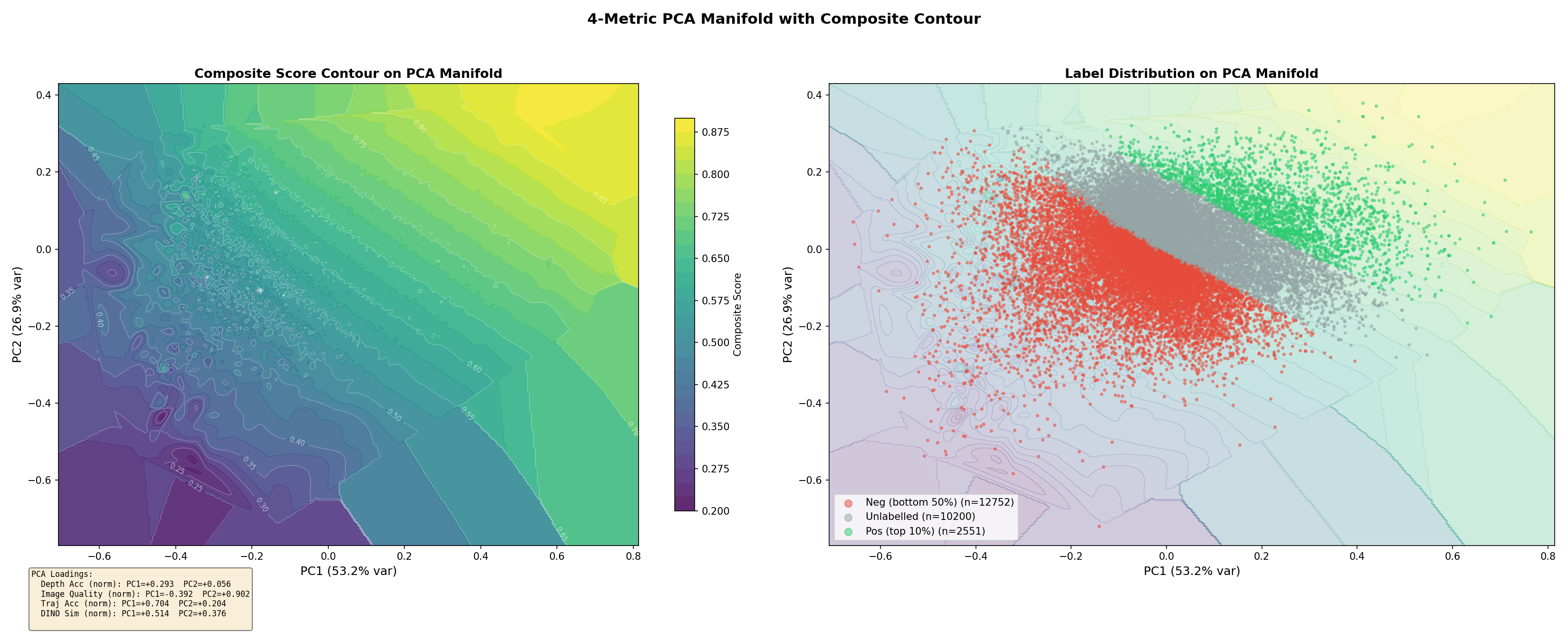}
  \caption{Two-dimensional PCA manifold of the four metrics: left, composite score contours; right, label scatter overlaid on faded contours. PC1 (53.2\% variance) is driven mainly by trajectory accuracy and DINO similarity; PC2 (26.9\%) by image quality and depth accuracy. Composite increases diagonally, with Pos samples concentrated in the upper-right and Neg in the lower-left, showing that overall quality is geometrically separable in metric space and suitable as continuous supervision for the hallucination-aware model.}
  \label{fig:hmcase}
\end{figure}

\subsection{Baseline}
Our method is an extension of WoVR, which serves as our baseline. Rather than altering the world model’s structure, our goal is to unlock its existing potential with minimal overhead, improving training efficiency through our novel components.
We inherit the Action-Conditioned World Model from WoVR, using Wan2.2 as the backbone and injecting an action MLP to generate future image chunks. We do not include WoVR’s PACE strategy in our setup, as its improvements are orthogonal to our method: any performance boost from PACE would apply equally to our pipeline.
\subsection{Real-World Experiments}

\begin{figure}[!htbp]
    \centering
    \includegraphics[width=0.85\textwidth]
        { 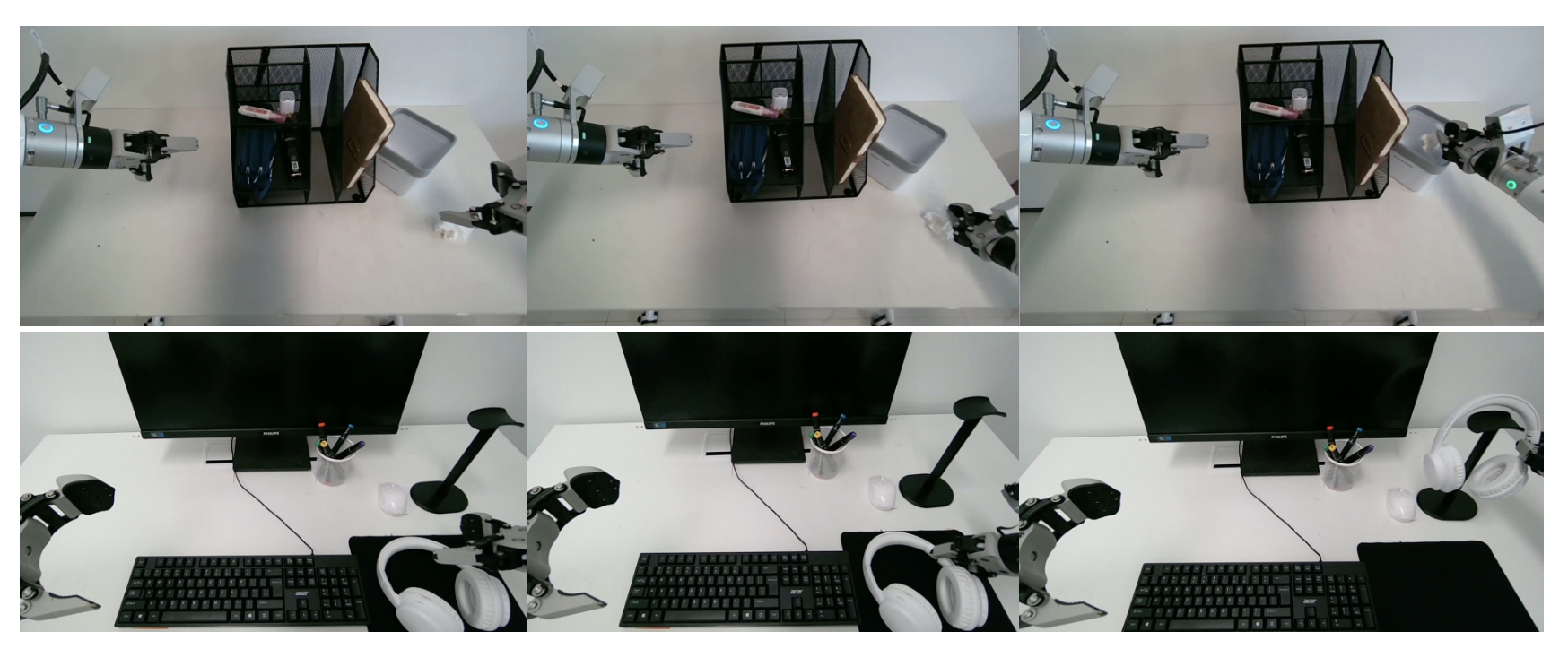}
    \caption{Real-world robot executions recorded during
    physical testing on the G1 robot.
    The upper row shows the Tissue-to-Box task, and the
    lower row shows the Headphone-on-Stand task.
    These images depict actual robot executions,
    not world-model-generated rollouts; aggregate
    success rates are reported in Table~\ref{tab:real_world}.}
    \label{fig:realworld_execution}
\end{figure}

We evaluate HaWMPO on the G1 robot using two physical
manipulation tasks: \textit{Tissue-to-Box}, in which the robot
places a tissue into a white storage box, and
\textit{Headphone-on-Stand}, in which it places headphones
onto a stand. Fig.~\ref{fig:realworld_execution} shows
actual robot executions recorded during physical testing,
rather than world-model-generated predictions.

We compare HaWMPO with the SFT-trained OpenVLA-OFT base
policy and WoVR under the same evaluation protocol,
with 20 trials per method per task. A trial is successful
if the tissue remains inside the box or the headphones
remain supported by the stand after release.
On Tissue-to-Box, the base policy, WoVR, and HaWMPO
achieve success rates of 75\% (15/20), 80\% (16/20),
and 85\% (17/20), respectively. On Headphone-on-Stand,
the corresponding success rates are 60\% (12/20),
65\% (13/20), and 75\% (15/20).

Across the two tasks, HaWMPO achieves an average success
rate of 80.0\%, compared with 67.5\% for the SFT baseline
and 72.5\% for WoVR, yielding improvements of 12.5 and
7.5 percentage points, respectively. These results provide
preliminary evidence of benefits in physical manipulation,
although evaluation on additional tasks and trials is
needed to establish broader generalization.

\subsection{World Model as Simulator}
\label{sec:wm_simulator}

\paragraph{Model architecture.}
We use an action-conditioned adaptation of Wan2.2-TI2V-5B
as the world-model simulator. The model consists of a video
variational autoencoder (VAE) and a latent-space diffusion
Transformer (DiT). The VAE encodes visual observations into
spatiotemporal latents and decodes predicted latents into RGB
frames. The DiT contains 30 Transformer blocks with a hidden
dimension of 3072 and 24 attention heads. Each block combines
self-attention over video tokens, cross-attention to conditioning
tokens, and a feed-forward network.

Robot actions are incorporated through two conditioning paths.
First, an MLP projects individual actions into tokens that
serve as keys and values for cross-attention.
Second, temporally grouped actions are projected into
embeddings that are added to the diffusion-timestep embeddings,
thereby conditioning the modulation of the Transformer blocks.
This design conditions visual predictions on the commanded
motion rather than relying on visual context alone.

\paragraph{Inputs and outputs.}
Each simulator step receives five conditioning RGB frames:
a fixed reference frame and the four most recent frames.
It additionally receives an eight-step action chunk,
together with the action context associated with the
conditioning frames. In LIBERO, each action is seven-dimensional,
comprising three translational components, three rotational
components, and one gripper command.
All frames have a resolution of $256 \times 256$.

Let $\mathcal{C}_t$ denote the five-frame visual context,
$\mathcal{U}_t$ the associated action context, and
$A_t=(a_t,\ldots,a_{t+K-1})$ the proposed action chunk.
The world model predicts
\begin{equation}
    \widehat{O}_{t+1:t+K}
    \sim p_{\phi}\!\left(
        \,\cdot \mid \mathcal{C}_t,\mathcal{U}_t,A_t
    \right),
    \qquad K=8,
    \label{eq:wm_transition}
\end{equation}
where $\phi$ denotes the world-model parameters and
$\widehat{O}_{t+1:t+K}$ contains eight predicted RGB frames.
The generation pipeline decodes a 13-frame sequence,
including the five conditioning frames; only the eight
future frames are returned as the simulated transition.
The conditioning latents are held fixed during iterative
denoising. The simulator is conditioned on images and actions;
task instructions are supplied to the VLA policy rather than
directly to the video-generation model in this implementation.

\paragraph{Closed-loop imagined rollouts.}
A rollout is initialized from recorded visual context.
The VLA policy predicts an action chunk from the current
observation and task instruction, and the world model
generates the corresponding future observations.
The last four generated frames then update the visual context,
while the reference frame remains fixed.
Repeating this process produces a closed-loop imagined
trajectory. Thus, generation is autoregressive across chunks,
whereas each future chunk is generated jointly through
latent-space denoising.

The world model is trained before policy optimization and
remains frozen during reinforcement learning.
It predicts visual transitions, not rewards or hallucination
scores: a separate reward model evaluates the generated
observations, and HAM estimates their reliability.
Reward-Soft uses the HAM scores to modulate the rewards
used for policy optimization.

\begin{figure}[!htbp]
    \centering
    \includegraphics[width=0.95\linewidth]{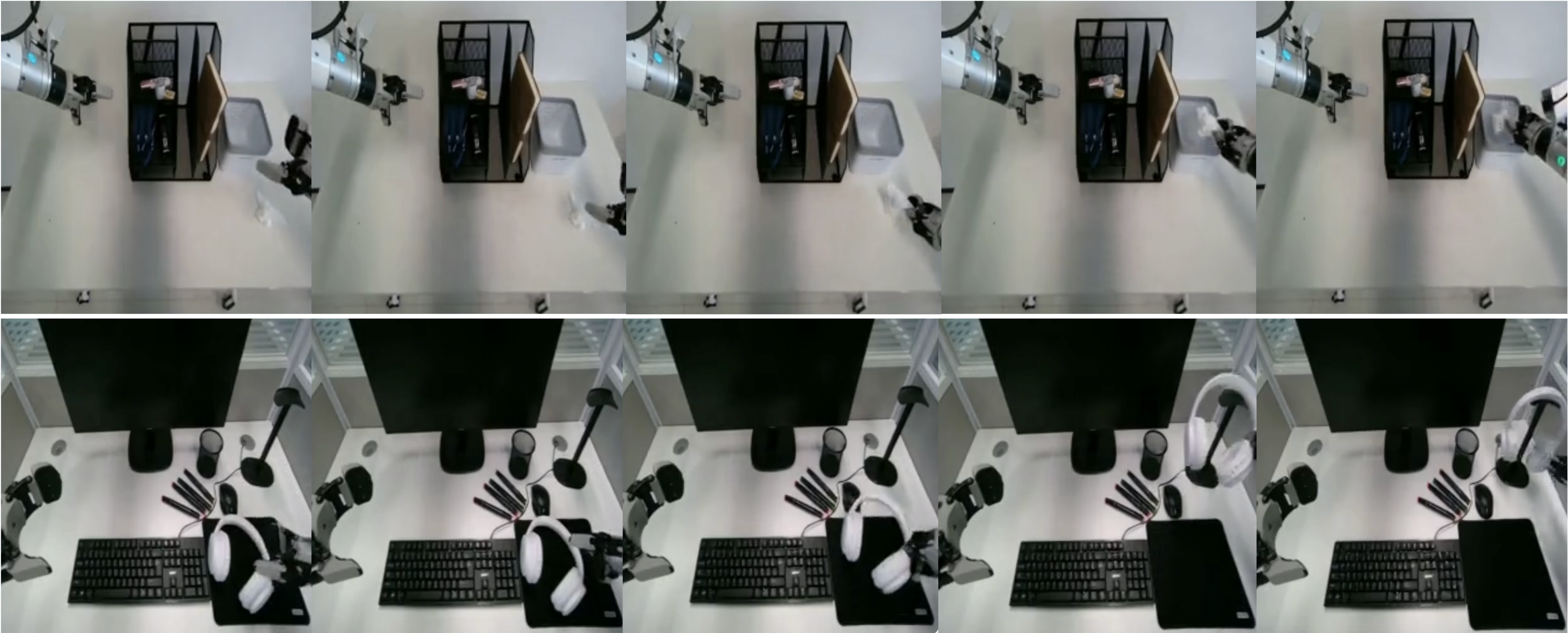}
    \caption{Selected world-model rollout frames for
    Tissue-to-Box (top), placing a tissue into a white
    storage box, and Headphone-on-Stand (bottom), placing
    headphones onto a stand. Frames are ordered
    chronologically from left to right. These examples
    illustrate action-conditioned visual predictions
    in physical manipulation scenes.}
    \label{fig:wm_simulator}
\end{figure}

Fig.~\ref{fig:wm_simulator} presents qualitative examples
of the generated rollouts. Although the predictions depict
task-relevant changes in the scene, visual plausibility alone
does not guarantee accurate contact dynamics or successful
physical execution. Errors can accumulate as generated
observations are reused as context, motivating the
hallucination-aware reward modulation in HaWMPO.
\end{document}